\documentclass[sigconf,screen,nonacm]{acmart} %nonacm

\usepackage{amsmath}
\usepackage[utf8]{inputenc} % allow utf-8 input
\usepackage[T1]{fontenc}    % use 8-bit T1 fonts
\usepackage{hyperref}       % hyperlinks
\usepackage{url}            % simple URL typesetting
\usepackage{booktabs}       % professional-quality tables
\usepackage{amsfonts}       % blackboard math symbols
\usepackage{nicefrac}       % compact symbols for 1/2, etc.
\usepackage{microtype}      % microtypography
\usepackage{xcolor}         % colors

\usepackage{xcolor,soul}
\sethlcolor{yellow}

\usepackage{tabularx, booktabs} % in preamble
\newcolumntype{Y}{>{\centering\arraybackslash}X}

\AtBeginDocument{%
  }

\setcopyright{acmlicensed}
\copyrightyear{2026}
\acmYear{2026}
\setcopyright{cc}
\setcctype{by}
\acmConference[FSE Companion '26]{34th ACM Joint European Software Engineering Conference and Symposium on the Foundations of Software Engineering}{July 05--09, 2026}{Montreal, QC, Canada}
\acmBooktitle{34th ACM Joint European Software Engineering Conference and Symposium on the Foundations of Software Engineering (FSE Companion '26), July 05--09, 2026, Montreal, QC, Canada}
\acmDOI{10.1145/3803437.3805232}
\acmISBN{979-8-4007-2636-1/2026/07}

\begin{document}

%%
%% The "title" command has an optional parameter,
%% allowing the author to define a "short title" to be used in page headers.
\title[Graph Embedding-Based Anomaly Detection in Microservice Architectures]{From Load Tests to Live Streams: Graph Embedding-Based Anomaly Detection in Microservice Architectures}

%%
%% The "author" command and its associated commands are used to define
%% the authors and their affiliations.
%% Of note is the shared affiliation of the first two authors, and the
%% "authornote" and "authornotemark" commands
%% used to denote shared contribution to the research.
\author{Srinidhi Madabhushi}
% \email{srimadab@amazon.com}
\affiliation{%
  \institution{Amazon Prime Video}
  % \city{Seattle}
  % \state{Washington}
  \country{USA}}

\author{Pranesh Vyas}
%\email{vyaspran@amazon.com}

\affiliation{%
  \institution{Amazon Prime Video}
  % \city{Austin}
  % \state{Texas}
  \country{USA}}

\author{Swathi Vaidyanathan}
% \email{vaiswath@amazon.com}
\affiliation{%
  \institution{Amazon Prime Video}
  % \city{Seattle}
  % \state{Washington}
  \country{USA}}

\author{Mayur Kurup}
% \email{kurupm@amazon.com}

\affiliation{%
  \institution{Amazon Prime Video}
  % \city{Austin}
  % \state{Texas}
  \country{USA}}

\author{Elliott Nash}
% \email{enash@atlassian.com}

\affiliation{%
  \institution{Amazon Prime Video}
  % \city{Seattle}
  % \state{Washington}
  \country{USA}}

  \author{Yegor Silyutin}
% \email{silyutin@amazon.com}

\affiliation{%
  \institution{Amazon Prime Video}
  % \city{Seattle}
  % \state{Washington}
  \country{USA}}

\begin{abstract}

Prime Video regularly conducts load tests to simulate the viewer traffic spikes seen during live events such as \textit{Thursday Night Football} as well as video-on-demand (VOD) events such as \textit{Rings of Power}. While these stress tests validate system capacity, they can sometimes miss service behaviors unique to real event traffic. We present a graph-based anomaly detection system that identifies under-represented services using unsupervised node-level graph embeddings. Built on a GCN-GAE, our approach learns structural representations from directed, weighted service graphs at minute-level resolution and flags anomalies based on cosine similarity between load test and event embeddings. The system identifies incident-related services that are documented and demonstrates early detection capability. We also introduce a preliminary synthetic anomaly injection framework for controlled evaluation that show promising \textbf{precision (96\%)} and low \textbf{false positive rate (0.08\%)}, though \textbf{recall (58\%)} remains limited under conservative propagation assumptions. This framework demonstrates practical utility within Prime Video while also surfacing methodological lessons and directions, providing a foundation for broader application across  microservice ecosystems.

\end{abstract}

\begin{CCSXML}
<ccs2012>
   <concept>
<concept_id>10010147.10010257.10010258.10010260.10010229</concept_id>
       <concept_desc>Computing methodologies~Anomaly detection</concept_desc>
       <concept_significance>500</concept_significance>
       </concept>
 </ccs2012>
\end{CCSXML}

\ccsdesc[500]{Computing methodologies~Anomaly detection}
%%
%% Keywords. The author(s) should pick words that accurately describe
%% the work being presented. Separate the keywords with commas.
\keywords{Graph Representation, Representation Learning, Anomaly Detection, Microservice Architecture, Graph Neural Network}

% \received{20 February 2007}
% \received[revised]{12 March 2009}
% \received[accepted]{5 June 2009}

%%
%% This command processes the author and affiliation and title
%% information and builds the first part of the formatted document.
\maketitle

%%% -------------------------------------------------
%%% 1 INTRODUCTION
%%% -------------------------------------------------

\section{Introduction}

Modern cloud-based applications such as Prime Video are powered by complex microservice architectures, where hundreds of specialized services interact to deliver seamless customer experiences. These distributed systems generate massive volumes of interaction data, especially during high-velocity events like Thursday Night Football (TNF), Champions League (UCL) and other peak usage periods. The resulting service-to-service communications form evolving graph streams that encode both the structural dependencies between services and the temporal dynamics of their interactions.  

To prepare for such events, Prime Video runs “gameday” load tests that stress the microservice stack weeks ahead of time. These tests validate scaling and infrastructure resilience under peak load~\cite{aws-gameday}. However, generating truly representative traffic patterns at scale remains difficult. Gamedays often exhibit anomalous load distributions and resource inefficiencies that diverge from real-world customer behavior. In live operations, incident response teams must identify and mitigate service-level issues triggered by faulty deployments, configuration changes, or resource bottlenecks, often with limited contextual awareness of dependencies. The lack of systemic observability and the unrealistic nature of gameday traffic both create risks: reduced availability, performance degradation, wasted capacity, increased operational overhead, and delayed incident detection—all of which directly affect customer experience and business continuity. To address this, we ask:
\begin{quote}
\textit{How can we use unsupervised graph-based representation learning to detect nodes whose behavior on gameday deviates from actual event day traffic?}
\end{quote}

Two key gaps motivate our approach:

\begin{enumerate}
  \item \textbf{Service-mix skew:} Gamedays address peak volume that is expected but miss per-service interaction patterns, causing over- or under-testing.
  \item \textbf{Hidden dependency cascades:} Small upstream changes may propagate silently, degrading downstream performance without breaching Service Level Agreement (SLA).
\end{enumerate}

At Prime Video’s scale, where a single action may touch hundreds of services, manual validation becomes time consuming. To address this, we propose an unsupervised graph-based anomaly detection system based on structural representations of minute-level service interaction graphs by adopting Graph Convolutional Autoencoders (GCN-GAE)~\cite{kipf2016variational}. Our key contributions are as follows:
\begin{enumerate}
\item \textbf{Multi-Snapshot Training.} We extend GCN-GAE to train across independently sampled, weighted graph snapshots, enabling scalable learning without temporal dependencies.
\item \textbf{Anomaly Scoring and Evaluation.} We use cosine similarity for node-level anomaly scoring and validate performance via documented incidents and a synthetic injection framework.
\item \textbf{Operational Insights.} We study detection lead time, false positives, and explainability needs in production, and suggest how contextual signals can enhance interpretability.
\end{enumerate}

Our system bridges the gap between simulated and real-world service behavior by focusing on structural graph deviations rather observing the raw \textbf{transactions per second (TPS)}. Beyond anomaly detection during gamedays, the system supports graph validation for structural shifts, change-safety gating through pre/post-deployment embedding drift analysis, and root-cause triage via neighborhood context to accelerate incident resolution. Together, these capabilities extend the utility of the detection system by providing service-level explainability, deployment safety validation, and anomaly awareness—all within a unified graph framework.

\section{System Context}

Prime Video's backend follows a two‐tiered microservice architecture spanning playback and storefront layers, having 200+ microservices~\cite{aws-reinvent}. This forms a deep, hierarchical service graph—starting sparsely at edge-facing services, becoming denser towards the middle of the graph, and thinning again at the end of the graph—with low clustering coefficients and minimal service/edge churn between graph snapshots. Each service uses demand-driven auto-scaling to maintain low latency and cost efficiency during high-traffic events like TNF and UCL. To model the complex service interactions, we represent system behavior as time-evolving service graphs. These capture both topological dynamics and fluctuating TPS across graph snapshots, as services and interactions emerge or disappear over time. 

\paragraph{Types of Graphs}

We define three graph types used in this work:
\begin{enumerate}
   \item \textbf{Baseline Graph:} Captured during steady-state, non-event periods; reflects normal traffic.
   \item \textbf{Event Graph:} Recorded during live and video-on-demand (VOD) events, representing real-world peak conditions.
   \item  \textbf{Gameday Graph:} Collected during an internal stress test to simulate event-like traffic at scale.
\end{enumerate}

\section{Problem Definition}

Let $\mathcal{G} = \{G_1, G_2, \dots, G_T\}$ be a sequence of dynamic graph snapshots from the microservice system. Each graph $G_t = (V_t, E_t, W_t)$ is a \textit{weighted, directed} graph where:
\begin{itemize}
  \item $V_t$ is the set of services active at time $t$;
  \item $E_t \subseteq V_t \times V_t$ represents directed inter-service calls;
  \item $W_t : E_t \rightarrow \mathbb{R}_{>0}$ is the TPS added as a weight to each edge $(u,v) \in E_t$.
\end{itemize}

\paragraph{GCN-GAE Architecture.} We adopt the graph autoencoder from~\cite{kipf2016variational}, using two GCN layers as the encoder and an inner-product decoder. The input to this model is the weighted adjacency matrix $A_t$ with dimension $|V_t|\times|V_t|$ and the feature matrix $X_t$ of dimension $|V_t|\times1$. We assume featureless input, i.e., $X_t = I$, an identity matrix. 

\paragraph{Structural Embedding.} For each graph $G_t$, we compute node embeddings $Z_t \in \mathbb{R}^{|V_t| \times d}$ with $d = 16 \ll |V_t|$, such that a decoder $\Psi$ can reconstruct the adjacency matrix $A_t$:
$\hat{A}_t = \Psi(Z_t) \approx A_t$.

\paragraph{Training Objective.} Model parameters are optimized by minimizing the reconstruction loss over a training subset $\mathcal{G}_{\text{train}} \subset \mathcal{G}$:
$\mathcal{L}_{\text{rec}} = \sum_{G_t \in \mathcal{G}_{\text{train}}} \| A_t - \hat{A}_t \|_F^2$.

\paragraph{Anomaly Scoring.} After training, we embed both gameday and reference event snapshots. For a service $v_i$, we compute cosine similarity:
$s_i = \frac{\langle Z_{\text{gameday}, i}, Z_{\text{event}, i} \rangle}
           {\| Z_{\text{gameday}, i} \|_2 \cdot \| Z_{\text{event}, i} \|_2}$.
A service is flagged as anomalous if $s_i < \tau$, for a given threshold $\tau \in (0,1)$.

\paragraph{Goal.} We aim to show that GCN-GAE embeddings trained using $\mathcal{L}_{\text{rec}}$ can reveal structural anomalies in production service graphs, without labeled incidents or temporal supervision.

%%% -------------------------------------------------
%%% 5  METHODOLOGY
%%% -------------------------------------------------

\section{Methodology}
\label{sec:challenge}

\begin{figure}[tb]
\centering
\includegraphics[width=\linewidth]{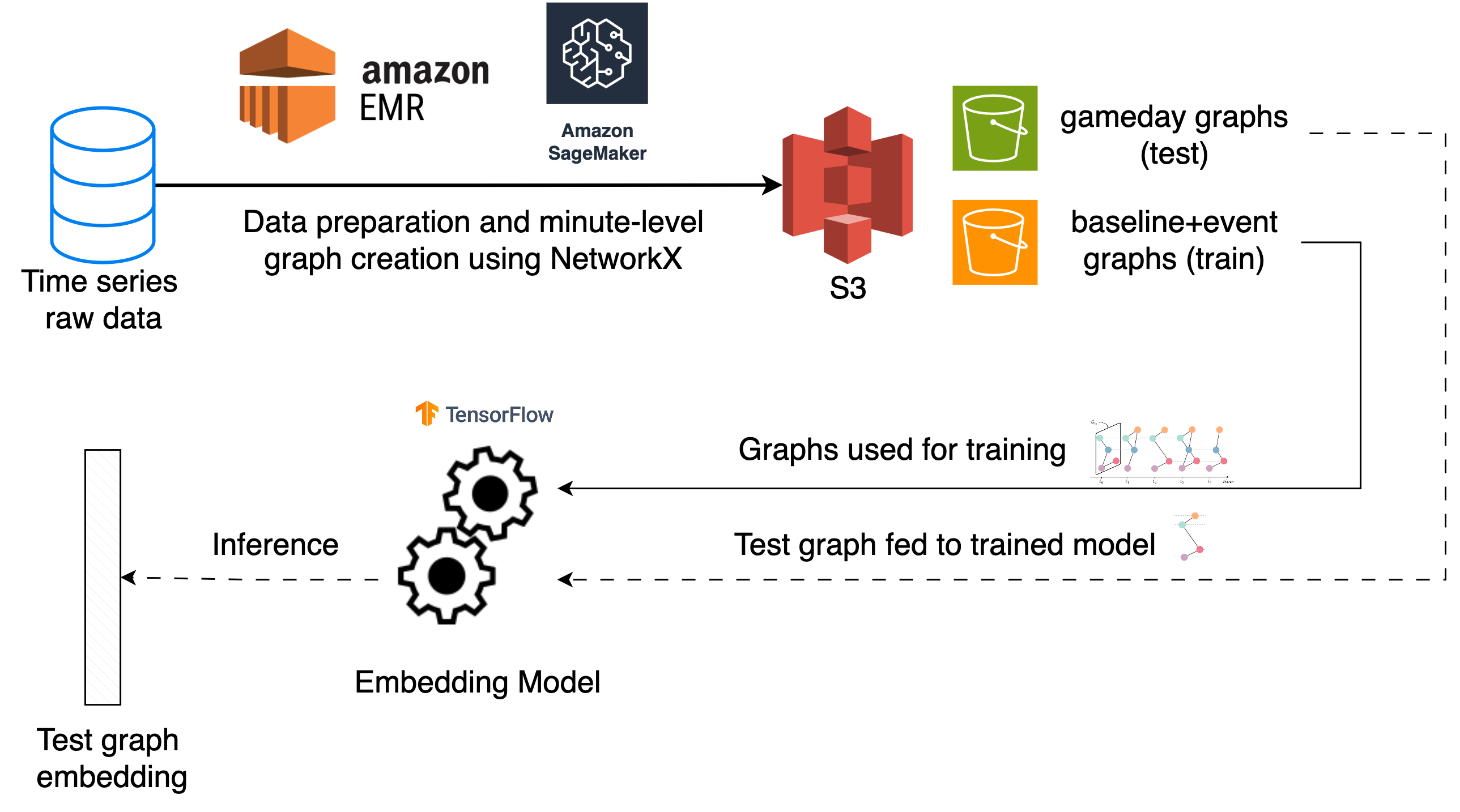}
\caption{End-to-end pipeline for training and evaluating service-level anomaly detection using graph embeddings.}
\label{fig:methodology-pipeline}
\end{figure}

Our end-to-end anomaly detection system is cloud-native and scalable, as shown in Figure~\ref{fig:methodology-pipeline}. It consists of three main stages: data ingestion, model training, and inference.

\paragraph{Data Ingestion and Preprocessing}

We use proprietary industry dataset containing raw time series telemetry (timestamp, source, destination, TPS) from Prime Video microservice architecture and aggregate it into minute-level dynamic graphs using Amazon EMR and Sagemaker. Each graph snapshot is stored as a NetworkX object and uploaded to Amazon S3. We generate graphs for five months, categorized as baseline (normal), event (TNF), or gameday (load test) graphs. Training uses baseline and event graphs to learn normal behavior. A reserved subset of event graphs serves as reference during inference. Gameday graphs are used only for evaluation to test detection under simulated conditions.

\paragraph{Model Training}

We implement a GCN-GAE model in TensorFlow, trained on normalized adjacency matrices using Amazon SageMaker. While we explored tdGraphEmbed~\cite{beladev2020tdgraphembed} and DySAT~\cite{sankar2020dysat}, tdGraphEmbed lacks node-level anomaly detection and DySAT incurs time-window dependencies with slower training and inference. We selected GCN-GAE for its adaptability to dynamic graphs and traffic scales, combined with lightweight, fast inference. Monthly models are trained on baseline and event graphs over 50 epochs using the Adam optimizer ($\text{lr} = 10^{-2} $). Each model training completes in under 6 hours on an \texttt{ml.g5.2xlarge} instance.

We extend the GCN-GAE from~\cite{kipf2016variational} with key adaptations:
\begin{itemize}
  \item \textbf{Batching across graph snapshots:} We modified the code that handles only static graphs, to handle batch-wise training, allowing the model to generalize across diverse baseline and event patterns.
  \item \textbf{Loss function modification:} Since the original model assumes binary edges for link prediction task, we replaced binary cross-entropy with mean squared error to better reflect the reconstruction accuracy of weighted edges.
  \item \textbf{Optimizer update logic:} The optimization process was updated to accumulate loss over a batch and apply parameter updates at the end of each batch, in contrast to the original setup which was designed for static graph updates.
\end{itemize}

\paragraph{Inference and Similarity Analysis}

At inference, gameday graphs are encoded into node embeddings and compared against reference event embeddings via cosine similarity. Services scoring below 0.98 are flagged as anomalous. This threshold was empirically calibrated from historical baseline and event graphs, where scores were consistently above 0.98, providing a reliable boundary for detecting structural deviations while minimizing false positives.

\section{Evaluation Results}
\label{sec:evaluation}

The analysis in this section is organized around five research questions (RQs) that guide our evaluation strategy. We compare graphs using cosine similarity, validate findings with documented incident tickets, and inject synthetic anomalies for benchmarking.

\subsection{Embedding Quality Evaluation}
\paragraph{\textbf{RQ1.} Do the generated embeddings effectively capture general Prime Video graph behavior with respect to quality?}

\begin{figure}[t]
  \centering
  \includegraphics[width=0.75\columnwidth]{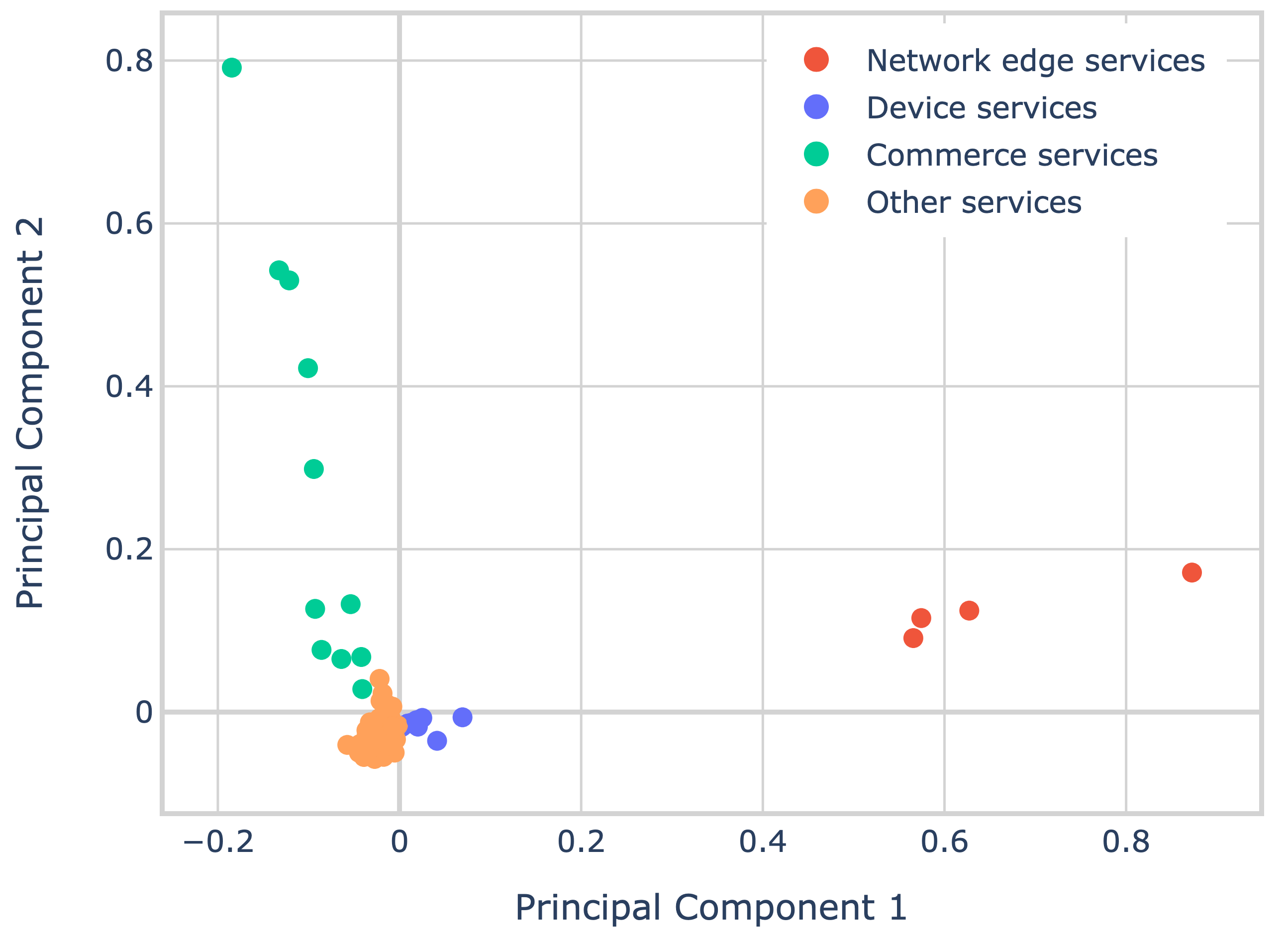}
  \caption{PCA over service embeddings showing natural groupings and neighborhood structure.}
  \label{fig:embedding-quality-clustering}
\end{figure}

We assess embedding quality via reconstruction loss across baseline, event, and gameday graphs (Table~\ref{tbl:recon_loss}). Loss remains low for baseline and event graphs as the models were trained on similar profiles, while gameday shows elevated loss as such examples were omitted during training, confirming structural differences from normal customer traffic. PCA visualization (Figure~\ref{fig:embedding-quality-clustering}) also shows that embeddings cluster well, grouping services with similar functionality together. Network edge services receive customer traffic at the first layer, device services handle various device types (TVs, phones), commerce services manage transactions and subscriptions, and the remaining services handle app requests and deliver streaming content~\cite{aws-reinvent}.

\begin{table}[t]
\centering
\caption{Reconstruction losses for different graph types.}
\label{tbl:recon_loss}
\begin{tabular}{lccc}
\toprule
\textbf{Graph Type} & \textbf{Min Loss} & \textbf{Median Loss} & \textbf{Max Loss} \\
\midrule
Baseline   & 0.0008 & 0.0011 & 0.0013 \\
Event      & 0.0009 & 0.0012 & 0.0014 \\
Gameday    & 0.0942  & 0.1567  & 0.2051  \\
\bottomrule
\end{tabular}
\end{table}

\subsection{Ground Truth Evaluation}

A \textit{Correction of Error} (CoE) is an Amazon mechanism for documenting and analyzing incidents via structured root cause analysis~\cite{aws-coe}. Each CoE includes information such as impacted services, customer impact, root cause, security implications, and corrective actions.

\paragraph{\textbf{RQ2.} Does the model successfully detect all services referenced in CoEs?}

To answer this, we analyze three CoEs provided by stakeholders and evaluate whether our method identifies the services listed in each report. Table~\ref{tbl:coe-results} summarizes the number of services mentioned and how many were detected by our system. In two cases, the model successfully identified all referenced services, detecting them 3 minutes before the first alarm was raised. However, CoE \#3 revealed a limitation: the impact propagated beyond the GCN layers' receptive field to an unrelated service outside the affected call path. Capturing such long-range effects would require deeper models, which risk oversmoothing~\cite{rusch2023survey}, suggesting the need for rule-based or complementary modeling as future work.

\begin{table}[h]
    \caption{Anomaly detection accuracy compared to documented service incidents in CoEs.}
    \label{tbl:coe-results}
    \begin{tabularx}{\columnwidth}{cXc}
        \toprule
        \textbf{CoE}  & \textbf{Description} & \textbf{Detected} \\
        \midrule
        \#1 & Outage affected viewers during an event due to a service bug that activates only during live broadcasts & 1 / 1 \\
        \#2 & Gameday was terminated early when service overload caused errors for customers attempting to view video details & 2 / 2 \\
        \#3 & Customers couldn't play content because a new feature in one service unexpectedly affected an unrelated service & 0 / 1 \\
        \bottomrule
    \end{tabularx}  
\end{table}

\subsection{Explaining Detected Anomalies}

Given the unsupervised nature of our setup, the model often flagged anomalous services beyond those explicitly linked to known CoE incidents. Validating these at scale required a systematic approach, motivating the exploration of additional validation sources, addressed in the following research question.

\paragraph{\textbf{RQ3.} What communication channels are typically required to explain identified anomalies?}

To answer this, we explored several avenues for anomaly validation and explanation.

\textbf{Source 1: Follow-up Analysis.} Our initial method involved analyzing \textit{fan-out ratios}—the proportion of TPS for a service's single outgoing edge to the total incoming TPS of that service that span over all its incoming edges. Although total traffic volume varied across graphs, fan-out ratios remained relatively stable under normal conditions. This enabled us to use deviations as a lightweight diagnostic for anomalous behavior. In several cases, abrupt changes in fan-out ratios explained anomalies. Table~\ref{tbl:fanout} shows three upstream services of an anomalous service. Though undocumented in CoEs and below incident severity, fan-out ratio deviations validated the model's alert: two upstream services exhibited significantly higher percentage differences between normal and anomalous graphs.

\begin{table}[h]
   \caption{Absolute percentage difference in fan-out ratios for the top 3 upstream edges of an anomalous service.}

    \label{tbl:fanout}
    \begin{tabularx}{\columnwidth}{XYYY}
        \toprule
        \textbf{Upstream service} & \textbf{\% incoming traffic} & \textbf{\% fan-out diff. normal vs. normal} & \textbf{\% fan-out diff. normal vs. anomaly}  \\

        \midrule
       Service 1    & 61\% & 4.4\%  & 33.1\%  \\
        Service 2 & 21\% & 2.3\% & 115.7\% \\
        Service 3  &   18\% & 1.9\% & 4.1\% \\

        \bottomrule
    \end{tabularx}  
\end{table}

\begin{figure}[h]
  \centering
  \includegraphics[width=0.85\columnwidth]{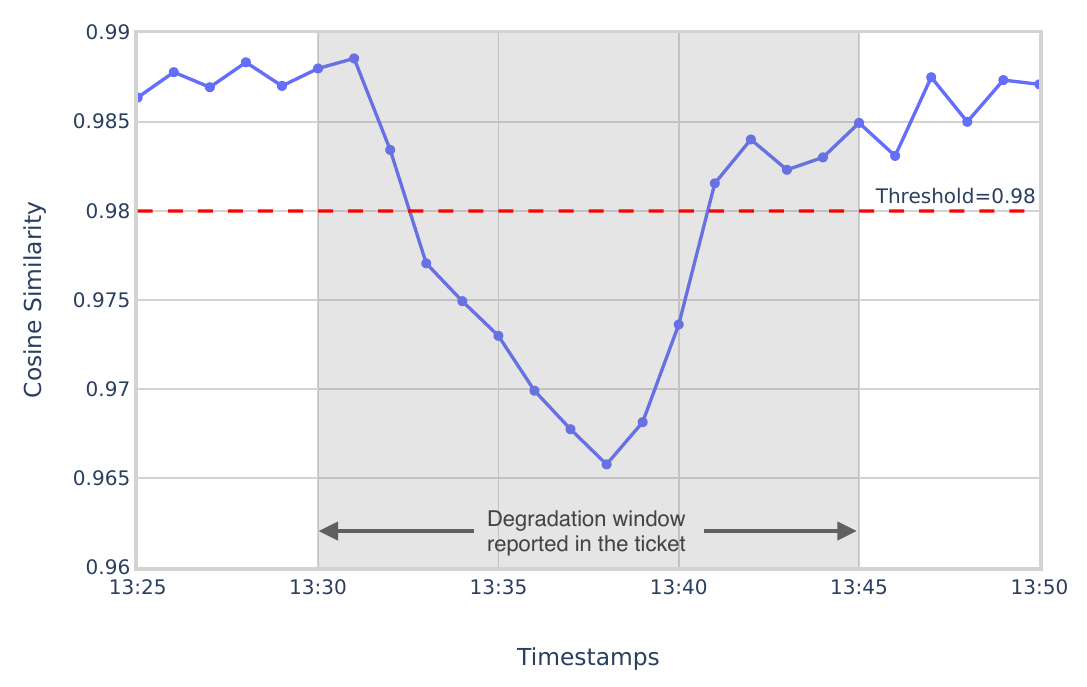}
  \caption{A brief dip in cosine similarity for a service over a ten-minute window, later corroborated by an RCA ticket.}
  \Description{The cosine similarity dip aligns with a one-hour incident from an RCA ticket, validating the model’s detection}
  \label{fig:explain-with-rca}
\end{figure}

\textbf{Source 2: RCA Tickets.} Root Cause Analysis (RCA) tickets document diagnoses and resolutions of failures in the microservice stack, often listing contributing factors and mitigations~\cite{aws-rca}. To validate anomalies unexplained by fan-out ratios, we reviewed a subset of RCA tickets that were created around the time of the anomaly. Figure~\ref{fig:explain-with-rca} shows one example: a service with a subtle cosine similarity dip over ten minutes was flagged. Though not in a CoE and showing no fan-out issues, an RCA ticket reported a 15-minute incident involving this service. Our model’s detection fell within that window, suggesting the anomaly was real and aligned with internal failures.

\begin{figure}[h]
  \centering
  \includegraphics[width=\columnwidth]{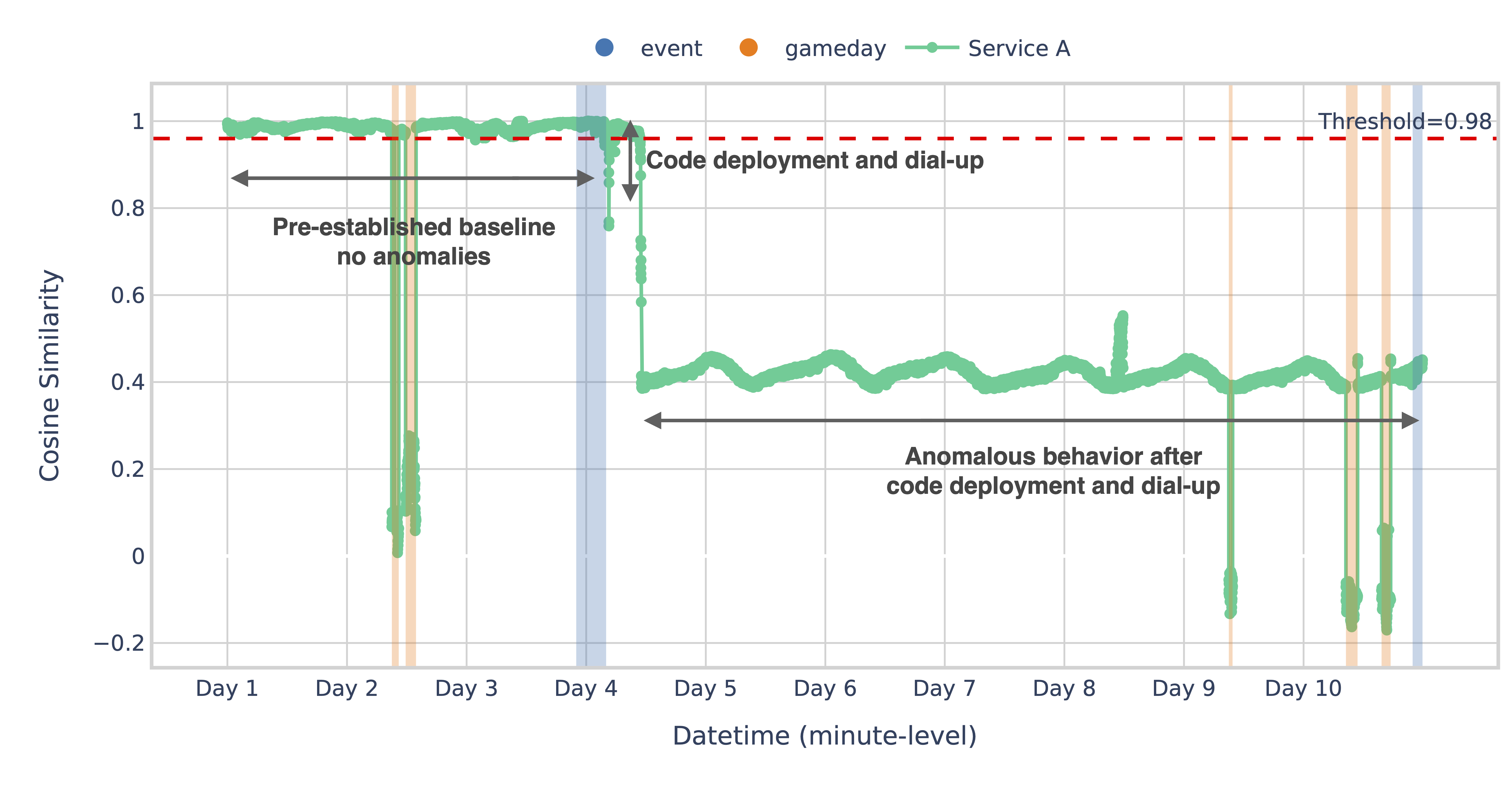}
  \caption{A shift in cosine similarity after a code deployment}
  \Description{Cosine similarity shows a clear shift in behavior following a code deployment}
  \label{fig:explain-with-spoc}
\end{figure}

\textbf{Source 3: Validating with a SPOC.} 
Figure~\ref{fig:explain-with-spoc} shows that gameday timestamps (vertical orange strips) were correctly flagged as anomalous due to traffic patterns unrepresentative of normal operations. However, cosine similarity scores also dropped persistently from Day 4 onward, prompting further investigation. At Prime Video, each service is owned by a team with a Single Point of Contact (SPOC)—a central communication route connecting questions to the service team~\cite{topdesk-spoc}. The service's SPOC confirmed a code deployment on Day 4 that altered traffic proportions. Since the model was trained only on historical patterns, this shift appeared novel. Daily seasonality remained stable, and no related services exhibited similar changes, confirming the effect was localized. This case demonstrates the framework's value for both gameday anomaly detection and change safety: detecting behavioral shifts and assessing their propagation can support safer rollouts and proactive validation.

\paragraph{\textbf{RQ4.} Which benign services are misclassified or remain unexplained by the model?}

\begin{figure}[ht]
  \centering
  \includegraphics[width=0.85\columnwidth]{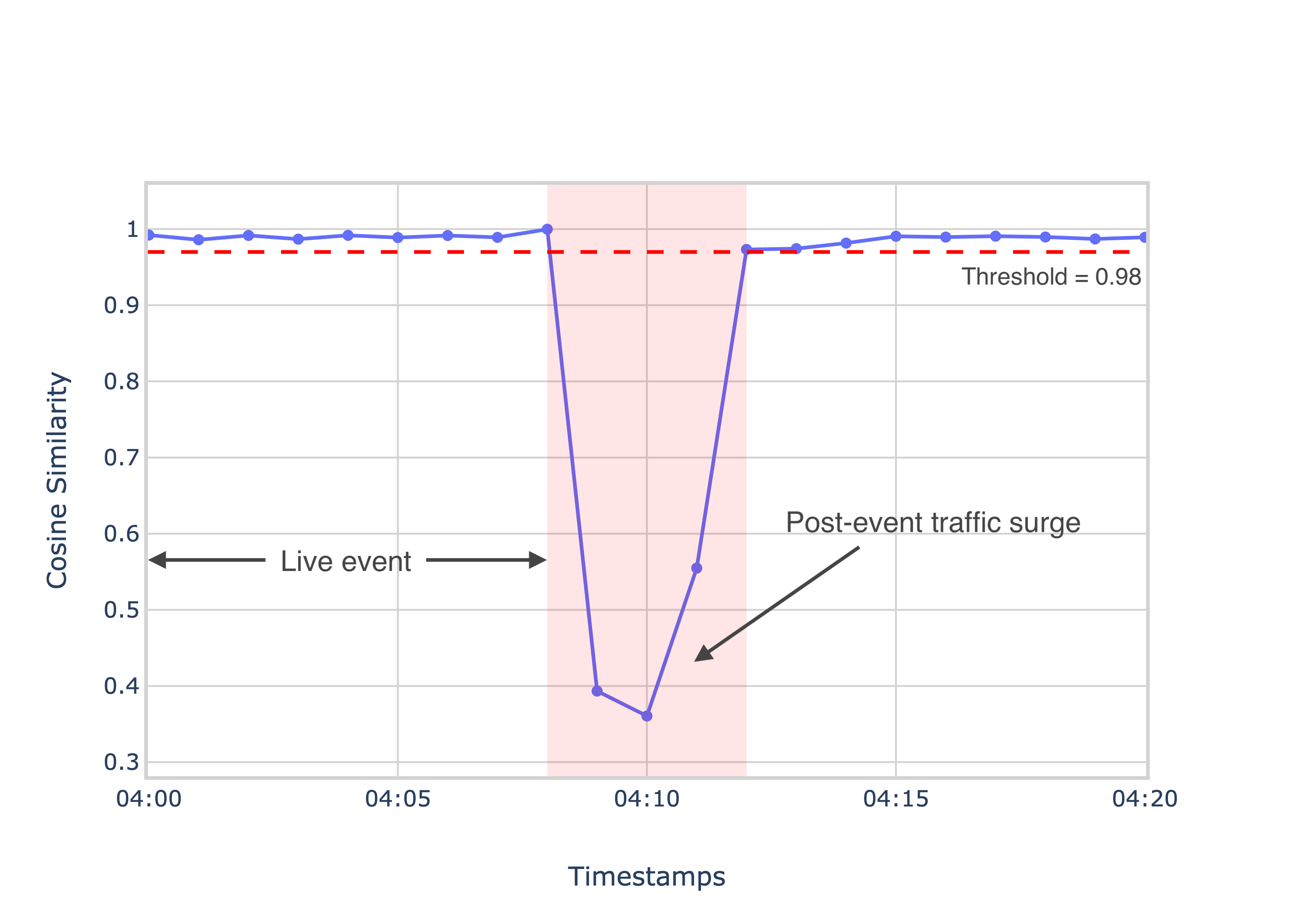}
  \caption{Cosine similarity of a homepage redirect service after a live event being incorrectly tagged as anomaly.}
  \Description{Cosine similarity shows a significant deviation during the post-event traffic surge, which was a legitimate but previously unseen pattern.}
  \label{fig:explain-with-context}
\end{figure}

We investigated cases where flagged anomalies remained unexplained by prior methods but were contextually reasonable. For example, Figure~\ref{fig:explain-with-context} shows a service flagged after a live event. No incident was recorded, but TPS spiked as the service redirected users to the homepage after a live event ended—resulting in a legitimate traffic surge. Other anomalous services that were detected lacked abnormal metrics or clear dependencies to true anomalies. This suggests the model may surface latent correlations beyond visible metadata. While promising, this complicates interpretability—enhancing explainability will be key to reducing false positives and building operator confidence.

\subsection{Synthetic Anomaly Injection Evaluation} 

While our anomaly detection system passed several sanity checks and identified services involved in past incidents, a more rigorous evaluation framework was needed to assess its performance in a controlled setting. As the model is tailored to a specific production environment, we did not benchmark against existing models, since meaningful comparisons would require extensive customization of the models. Instead, we evaluated the model in isolation using a synthetic anomaly injection framework, allowing controlled anomaly introduction and quantifiable performance measurement.

\paragraph{\textbf{RQ5.} Does introducing a synthetic load along a selected call path improve anomaly detection evaluation?}
Answering this required careful design, as injecting synthetic anomalies is inherently non-trivial. Naively adding noise can yield ambiguous results, especially when a single anomaly cascades through dependencies, obscuring its origin. For the initial experiment, we selected a critical call path frequently involved in Prime Video user interactions, comprising 5 services and 4 connecting edges to enable clear and interpretable evaluation. For each selected edge, TPS was increased by a random percentage. The source and destination nodes of these modified edges were labeled as ground truth anomalies, enabling direct mapping between injected and expected detections. Initial results show \textbf{96\% precision} and \textbf{0.08\% false positive rate}, and \textbf{58\% recall}, a limitation stemming from our conservative assumption of complete anomaly propagation to all connected services, as the services on both ends of an edge are considered as ground truth anomalies. 
Future work includes improving how we capture service dependencies and developing smarter perturbation strategies that go beyond targeting individual edge paths.

\section{Key Takeaways}

This section presents key insights and lessons learned from testing our graph-based anomaly detection system in a high-scale, microservice environment.

\begin{itemize}
    \item \textbf{Structural Aspect is Sufficient.} While temporal models like DySAT~\cite{sankar2020dysat} capture evolution across time, we found per-minute structural embeddings sufficient. Time-window dependence limited applicability in our setting.  

    \item \textbf{Early Detection Advantage.} Our method consistently surfaced anomalies 1–3 minutes before corresponding high severity tickets. Node embeddings’ sensitivity to structural shifts provided meaningful lead time for response.  

    \item \textbf{Context Improves Explainability.} Anomalies did not align with TPS volume or fan-out alone. Incorporating upstream changes, deployment metadata, and historical patterns improved interpretability.  

    \item \textbf{Prioritization Reduces Noise.} False positives are inevitable at scale. Incorporating service criticality, blast radius, and customer impact into evaluation helps triage alerts and direct attention where most needed.  

    \item \textbf{Synthetic Injection Framework Needed.} Manual checks sufficed early on, but robust evaluation requires principled synthetic anomaly injection. Simulated faults with dependency effects provide realistic benchmarks for reliability.  

    \item \textbf{Proportionality Enables Generalization.} Normalizing edge weights per snapshot preserved interaction ratios, enabling fair comparisons across fluctuating TPS volumes and improving robustness to workload variation.  

    \item \textbf{Broad Applicability.} Beyond game day anomaly detection, the framework supports change validation, service clustering, system visualization, stress testing, and daily monitoring, demonstrating versatility across microservice operations.  
\end{itemize}

%%% -------------------------------------------------
%%% 8  RELATED WORK
%%% -------------------------------------------------
\section{Related Work}
\label{sec:related}

Graph neural networks (GNNs) are foundational for modeling graph data. The introduction of Graph Convolutional Networks (GCNs)~\cite{kipf2016semi} enabled scalable semi-supervised learning. Extensions such as Graph Attention Networks (GATs)~\cite{velickovic2017graph} and spectral filtering~\cite{defferrard2016convolutional} have improved adaptability across domains. Graph autoencoders (GAEs)~\cite{kipf2016variational} further support unsupervised learning by reconstructing graph structure, which underpins our modeling approach. For anomaly detection, deep models like DOMINANT~\cite{ding2019deep} and ADDGraph~\cite{zheng2019addgraph} combine structural and attribute signals in dynamic or attributed networks. Ma et al.~\cite{ma2023survey} survey such methods, though most rely on supervision or assume regular graph patterns, limiting their use in our unsupervised, production-grade setting. Dynamic graphs—where topology evolves—have been modeled with architectures like DySAT~\cite{sankar2020dysat} and EvolveGCN~\cite{pareja2020evolvegcn}, which capture temporal dependencies through self-attention or recurrent updates. These models require aligned time sequences, making them less applicable for our use case comparing disjoint graph snapshots. In production systems, graph-based approaches such as correlation analysis~\cite{meng2020cross} have helped infer service dependencies and detect anomalies. Our work connects dynamic graph learning, unsupervised anomaly detection, and production validation. By comparing structural embeddings from simulated gameday and event graphs, we detect anomalies without labels or temporal alignment. To our knowledge, this is the first application of structural graph embeddings for test vs. event validation at hyperscale.

%%% -------------------------------------------------
%%% 9  CONCLUSION
%%% -------------------------------------------------
\section{Conclusion and Future Work}
\label{sec:conclusion}
This work presents a graph-based anomaly detection framework using unsupervised structural embeddings to surface service-level deviations during Prime Video’s gamedays. By extending a GCN-GAE model to train across dynamic graphs, we show that missed anomalies can be detected, without labeled data or time-series forecasting. The model successfully identifies all services noted in CoE reports and enables early detection of emerging issues, demonstrating the promise of structural graph learning for reliability at scale. For future work, we plan to explore causality-aware techniques for prioritizing alerts based on service impact, expand validation using richer ground truth, validate on public datasets with improved anomaly injection frameworks for generalizable benchmarking, and develop more interpretable embedding analysis tools. These efforts aim to build a more principled, explainable, and production-ready anomaly detection system.

\bibliographystyle{ACM-Reference-Format}
\bibliography{references}
\end{document}